\documentclass[procedia]{easychair}
\usepackage{amsmath}
\usepackage{mathpazo}
\usepackage{doc}
\usepackage{makeidx}
\usepackage{caption,subfig}
\usepackage{xcolor}
\usepackage{wrapfig}
\usepackage{cite}
\usepackage{graphicx}
\def\procediaConference{99th Conference on Topics of
  Superb Significance (COOL 2014)}

\title{Towards a Calculus of Echo State Networks}

\titlerunning{Towards a Calculus of Echo State Networks}

\author{
    Alireza Goudarzi\inst{1}
\and
    Darko Stefanovic\inst{1,2}
}

\institute{
  Department of Computer Science, 
  University of New Mexico,
  Albuquerque, New Mexico, USA\\
  \email{alirezag@cs.unm.edu}
  \and
  Center for Biomedical Engineering, 
  University of New Mexico,
  Albuquerque, New Mexico, USA\\
   \email{darko@cs.unm.edu}\\
 }

\authorrunning{Goudarzi and Stefanovic}

\begin{document}

\maketitle

\keywords{Reservoir computing, echo state networks, analytical training, Wiener filters, dynamics}

\begin{abstract}
  Reservoir computing is a recent trend in neural networks which uses the dynamical perturbations on the phase space of a system to compute a desired target function. We present how one can formulate an expectation of system performance in a simple class of reservoir computing called echo state networks. In contrast with previous theoretical frameworks, which only reveal an upper bound on the total memory in the system, we analytically calculate the entire memory curve as a function of the structure of the system and the properties of the input and the target function. We demonstrate the precision of our framework by validating its result for a wide range of system sizes and spectral radii. Our analytical calculation agrees with numerical simulations. To the best of our knowledge this work presents the first  exact analytical characterization of the  memory curve in echo state networks.
\end{abstract}


%
%

\section{Introduction}
\label{sect:introduction}

In this paper we report our preliminary results in building a framework for a mathematical study of reservoir computing (RC) architecture called the echo state network (ESN).
Reservoir computing (RC) is a recent approach in time series analysis that uses the perturbations in the intrinsic dynamics of a system, as opposed to its stable states, to compute a desired function. The classic example of reservoir computing is the echo state network, a recurrent neural network with random structure. These networks have shown good performance in many signal processing applications. The theory of echo state networks consists of analysis of memory capacity \cite{Dambre:2012fk}, e.g., how long can the network remember its inputs, and the echo state property \cite{DBLP:journals/nn/YildizJK12}, which consist of analysis of long term convergence of the phase space of the network. In RC, computation relies on the dynamics of the system and not its specific structure, which makes the approach an intriguing paradigm for computing with unconventional and neuromorphic architectures \cite{0957-4484-24-38-384004,0957-4484-18-36-365202,Snider:2005qa}. In this context, our vision is to develop special-purpose  computing devices that can be trained or ``programmed" to perform a specific task. Consequently, we would like to know the expected performance of a device with a given structure on given a task. Echo state networks (ESN) give us a simple model to study reservoir computing.  Extant studies of computational capacity and performance of ESN for various tasks have been carried out computationally and the main theoretical insight has been the upper bound for linear memory capacity \cite{Ganguli02122008,5629375,PhysRevLett.92.148102}.

Our aim is to use ESN to develop a theoretical framework that allows us to form an expectation about the performance of RC for a desired computation. To demonstrate the power of this framework, we apply it to the problem of memory curve characterization in ESN. Whereas previous attempts used the annealed approximation method to simplify the problem \cite{PhysRevLett.92.148102}, we derive an exact analytical solution to characterize the memory curve in the system. Our formulation reveals that ESN computes an  output as a linear combination of the correlation structure of the input signal and therefore the performance of ESN on a given task will depend on how well the output can be described as the input correlation in various time scales. Full development of the framework will allow us to extend our predictions to more complex tasks and more general RC architectures.

\section{Background}
\begin{figure}[ht]
\centering
\includegraphics[width=4in]{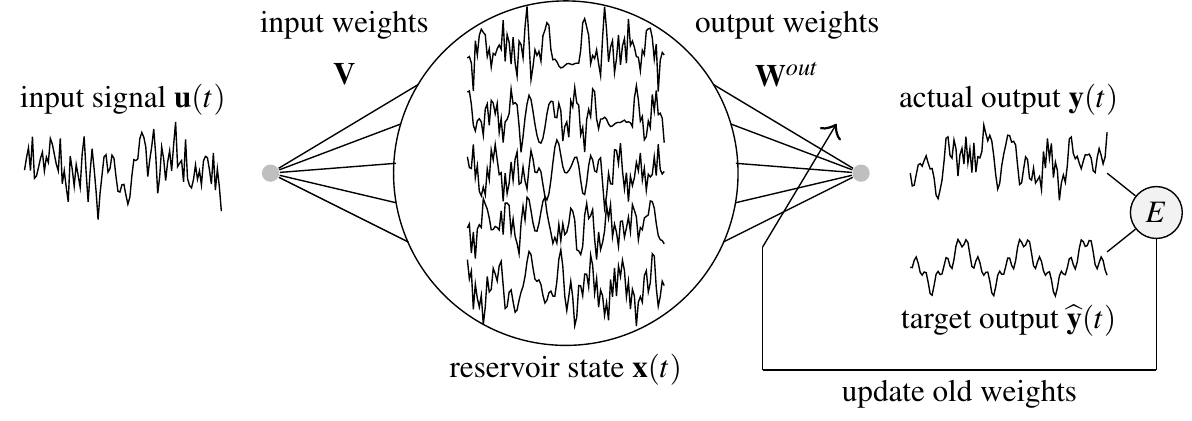}
\caption{Computation in a reservoir computer (RC). The reservoir is an excitable dynamical system with $N$ readable output states represented by the vector ${\bf X}(t)$. The input signal ${\bf u}(t)$ is fed
into one or more points $i$ in the reservoir with a corresponding weight $w^{in}_i$
denoted with weight column vector ${\bf W}^{in}=[w^{in}_i]$.}
\label{fig:dyn}
\end{figure}

In RC, a high-dimensional dynamical core called a {\em reservoir} is
perturbed with an external input. The reservoir states are then linearly combined to
create the output. The readout parameters can be calculated by performing regression
on the state of a teacher-driven reservoir and the expected teacher output. Figure~\ref{fig:dyn} shows an RC architecture. Unlike other forms
of neural computation, computation in RC
takes place within the transient dynamics of the reservoir.
The computational power of the reservoir is attributed to a short-term memory
created by the reservoir \protect\cite{Jaeger02042004} and the ability to preserve the
temporal information
from distinct signals over time \cite{Maass:2002p1444,Natschlaeger2003}. Several studies attributed this property to the dynamical regime of the reservoir and showed it to be optimal when the system operates in the critical dynamical regime---a regime
in which perturbations to the system's trajectory in its phase space neither spread nor
die out \protect\cite{Natschlaeger2003,Bertschinger:2004p1450,PhysRevE.87.042808,
4905041020100501,Boedecker2009}. The reason for this observation remains unknown.
 Maass et al. \protect\cite{Maass:2002p1444} proved that given the two
properties of {\em separation} and {\em approximation}, a reservoir system is capable
of approximating any time series. The separation property ensures that the reservoir
perturbations from distinct signals remain distinguishable, whereas the approximation
property ensures that the output layer can approximate any function of the reservoir
states to an arbitrary degree of accuracy. Jaeger \protect\cite{Jaeger:2002p1445} proposed that an ideal reservoir needs to have the so-called echo state property (ESP), which means that the reservoir
states asymptotically depend on the input and not the initial state of the reservoir. It
has also been suggested that the reservoir dynamics acts like a spatiotemporal
kernel, projecting the input signal onto a high-dimensional feature space \protect
\cite{Hermans:2011fk}.

\section{Model}
\label{sect:model}

\begin{figure}[b]
\centering
\includegraphics[width=3in]{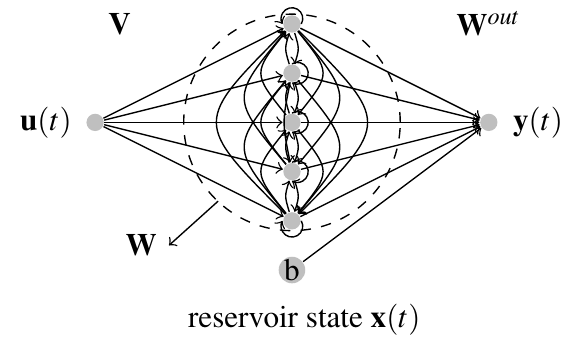}
\caption{Schematic of an echo state network (ESN). A dynamical core called a reservoir is driven by input signal ${\bf u}(t)$. The states of the reservoir ${\bf x}(t)$ extended by a constant $1$ and combined linearly to produce the output ${\bf y}(t)$. The reservoir consists of $N$ nodes interconnected with a random weight matrix ${\bf W}$. The connectivity between the input and the reservoir nodes is represented with a randomly generated weight matrix ${\bf W}^{in}$. The reservoir states and the constant are connected to the readout layer using the weight matrix ${\bf W}^{out}$. The reservoir and the input weights are fixed after initialization, while the output weights are learned using a regression technique.}
\label{fig:model}
\end{figure}
In this paper we restrict attention to linear ESNs, in which both the transfer function  of the reservoir nodes and the output layer are linear functions, Figure~\ref{fig:model}. The readout layer is usually a linear combination of the reservoir states. The readout weights are determined using supervised learning techniques, where the network is driven by a teacher input and its output is compared with a corresponding teacher output to estimate the error. Then, the weights can be calculated using any closed-form regression technique \cite{5629375} in offline training contexts.  Mathematically, the input-driven reservoir is defined as follows. Let $N$ be the size of the reservoir. We represent the time-dependent inputs as a column vector ${\bf u}(t)$, the reservoir state as a column vector ${\bf x}(t)$, and the output as a column vector ${\bf y}(t)$. The input connectivity is represented by the matrix ${\bf V}$ and the reservoir connectivity is represented by an $N\times N$ weight matrix ${\bf W}$. For simplicity, we assume  one input signal and one output, but the notation can be extended to multiple inputs and outputs. The time evolution of the linear reservoir is given by:

\begin{equation}
{\bf x}(t+1) = {\bf W}  {\bf x}(t) + {\bf V} {\bf u}(t),
\label{eq:timeevol}
\end{equation}
The output is generated by the multiplication of  an output weight matrix ${\bf W}^{out}$ of length  $N$ and the reservoir state vector $x(t)$:
\begin{equation}
{\bf y}(t) = {\bf W}^{out}{\bf x}(t).
\label{eq:output}
\end{equation}
The coefficient vector ${\bf W}^{out}$ is calculated to
minimize the squared output error $E=\langle ||{\bf y}(t)-{\bf \widehat{y}}(t)||^2 \rangle$ given the target output
${\bf \widehat{y}}(t)$. Here, $||\cdot||$ is the $L_2$ norm and $\langle\cdot\rangle$  the time average. The output weights are calculated using ordinary
linear regression using a pseudo-inverse form:

\begin{equation}
{\bf W}^{out} = \left\langle{\bf X}{\bf X'}\right\rangle^{-1}
\langle{\bf X} \widehat{\bf Y}'\rangle ,
\label{eq:regress}
\end{equation}
where each row $t$ in the matrix ${\bf X}$ corresponds to the state vector ${\bf x}(t)$, and ${\bf \widehat{Y}}$ is the target output matrix, whose rows correspond to target output vectors ${\bf \widehat{y}}(t)$.

\section{Deriving the Expected Performance}

Our goal is to form an expectation for the performance based on the structure of the reservoir and the properties of the task. Our approach is to calculate an expected output weight ${\bf W}^{out}$ using which the error of the system can be estimated. The calculation of ${\bf W}^{out}$ using the regression equation  Equation~\ref{eq:regress} has two components: the Gramm matrix $\left\langle{\bf X}{\bf X'}\right\rangle$, and the projection matrix ${\bf X}^T \widehat{\bf y}$. We now show how each of the two can be computed.

To compute the Gramm matrix, we start by expanding the recursion in Equation~\ref{eq:timeevol} to get an explicit expression for ${\bf x}(t)$ in terms of the initial condition and the input history:
\begin{align}
{\bf x}_t &=  {\bf W}^t {\bf x}_0 + \sum_{i=0}^{t-1}  {\bf W}^{t-i-1}  {\bf V}u_{i}, \\
\end{align}
Note that we have written the time as subscript for readability.
The first term in this expression is the contributions from the initial condition of the reservoir which will vanish for large $t$ when $\lambda^{max}<1$, where $\lambda^{max}$ is the spectral radius of the reservoir. Therefore  without loss of generality we can write:
\begin{align}
{\bf x}_t &=  \sum_{i=0}^{t-1}  {\bf W}^{t-i-1}  {\bf V}u_{i}, \\
\end{align}
Now we can expand the Gramm matrix as follows:

\begin{align}
\left\langle{\bf X}{\bf X'}\right\rangle &=  \frac{1}{T}\sum_{t=0}^{T}  {\bf x}_t\otimes {\bf x'}_t, \\
    &= \frac{1}{T}\sum_{t=0}^{T} \left(\sum_{i=0}^{t-1}{\bf W}^{t-i-1}  {\bf V}u_{i}\right)\otimes \left(\sum_{j=0}^{t-1}{\bf W}^{t-j-1}  {\bf V}u_{j}\right)'\\
    &= \frac{1}{T}\sum_{t=0}^{T} \left(\sum_{i=0}^{t-1} \sum_{j=0}^{t-1} {\bf W}^{t-i-1}  {\bf V}u_{i} \otimes {u'}_{j}{\bf V'}{\bf W'}^{t-j-1}  \right)
    \label{eq:xx}
\end{align}
The symbol $\otimes$ is the product. Similarly, we can write the projection matrix as:
\begin{align}
\langle{\bf X}\widehat{\bf Y}'\rangle &=  \frac{1}{T}\sum_{t=0}^{T}  {\bf x}_t \widehat{ y}'_t, \\
    &= \frac{1}{T}\sum_{t=0}^{T} \left(\sum_{i=0}^{t-1}{\bf W}^{t-i-1}  {\bf V}u_{i}\right)\otimes \widehat{ y}_t\\
    &= \frac{1}{T}\sum_{t=0}^{T} \sum_{i=0}^{t-1}{\bf W}^{t-i-1}  {\bf V}u_{i}\otimes \widehat{ y}_t'
    \label{eq:xy}
\end{align}
Note that in Equation~\ref{eq:xx} and Equation~\ref{eq:xy}, we have an explicit dependence on the reservoir weight matrix ${\bf W}$, the input weight matrix ${\bf V}$, the autocorrelation of input signal $u_i\otimes u_j$, and correlation of input and output $u_i\otimes \widehat{y}'_t$, at various time scales.  In the next section we use our fundamental equations, Equation~\ref{eq:xx} and \ref{eq:xy}, to calculate ${\bf W}^{out}$ and ultimately the expected performance of the system on a given task.

\section{Computing the Memory Curve}
In this section, we use our derivation to analytically calculate the memory capacity curve for a given reservoir structure defined by ${\bf W}$ and ${\bf V}$. Our assumption is that ${\bf W}$ is non-singular and has spectral radius $\lambda^{max}<1$. Memory capacity is the ability of the ESN to reconstruct its input after a delay of $\tau$. It is defined as~\cite{Jaeger:2001p1446}:
\begin{equation}
MC_{\tau} = \frac{\mathrm{Cov}^2(u_{t-\tau},y_t)}{\mathrm{Var}(u_t)\mathrm{Var}(y_t)},
\end{equation}
where $u_t$ is the input at time $t$, $u_{t-\tau}$ is the corresponding target output, and $y_t$ is the output of the network after calculating the ${\bf W}_{out}$. The inputs $u_t$ are drawn from identical and independent  uniform distributions in the range $[-1,1]$. The total memory capacity of a network is then given by:
\begin{equation}
MC = \sum_{\tau=1}^\infty MC_{\tau}.
\end{equation}
We now demonstrate how we can use the structure of the memory function and the i.i.d. property to analytically compute each component of ${\bf W}_{out}$.

To compute $\left\langle{\bf X}{\bf X'}\right\rangle$, we can assume that we have access to a one-dimensional infinitely long input-output stream to calculate the output weights. Then we can write:
\begin{align}
\left\langle{\bf X}{\bf X'}\right\rangle &= \lim_{T \to \infty}  \frac{1}{T}\sum_{t=0}^{T} \left(\sum_{i=0}^{t-1} \sum_{j=0}^{t-1} {\bf W}^{t-i-1}  {\bf V}u_{i} \otimes {u}_{j}{\bf V'}{\bf W'}^{t-j-1}  \right)\\
 &= \sum_{i=0}^{t-1} \sum_{j=0}^{t-1} \langle u_{i}  {u}_{j}\rangle  {\bf W}^{t-i-1}  {\bf V}\otimes{\bf V'}{\bf W'}^{t-j-1}
 \label{eq:xxmc}
\end{align}
Since $u_i$ are drawn from  i.i.d. uniform distribution between $[-1,1]$, we have:
\begin{align}
\langle u_{i}  {u}_{j}\rangle =
\begin{cases}
0,& \mbox{if } i\neq j\\
\frac{1}{3},& \mbox{if } i=j\\
\end{cases}
\end{align}
Moreover, let ${\bf W}={\bf UDU}^{-1}$ be the eigenvalue decomposition of $W$, ${\bf d}$ be a column vector corresponding to the  diagonal elements of ${\bf D}$,  $\bar{\bf V}={\bf U}^{-1}{\bf V}$ be the input weights in the basis defined by the eigenvectors of ${\bf W}$, and ${\bf I}^{\circ}$ the identity of Hadamard product denoted by $\circ$. We can rewrite Equation~\ref{eq:xxmc} as follows:
\begin{align}
\left\langle{\bf X}{\bf X'}\right\rangle &= {\bf U}\left(\lim_{t\to \infty}\sum_{i=0}^{t-1} \langle u_{i}  {u}_{i}\rangle \bar{\bf V}\otimes\bar{\bf V}'\circ {\bf d}^{i-1}\otimes {\bf d'}^{i-1} \right) {\bf U}'\\
&= \langle u^2\rangle{\bf U}\left(\bar{\bf V}\otimes\bar{\bf V}'\circ {\bf \bar{d}}\otimes{\bf \bar{d'}}\right){\bf U}'
\end{align}
Here $\langle u^2\rangle=\frac{1}{3}$ is the variance of the input, and ${\bf \bar{d}}$ is a column vector whose elements are $\bar{d}_l=\frac{1}{1-d_l}$. We have thus, computed the Gramm matrix as a function of the variance of the input,  ${\bf W}$, and ${\bf V}$. Note that for $MC_\tau$, $\left\langle{\bf X}{\bf X'}\right\rangle$ is constant for all $\tau$. However, for each $\tau$ we have to calculate  $\langle{\bf X}\widehat{\bf Y}'\rangle_\tau$ separately, denoted by the subscript $\tau$. Under the assumption of i.i.d. input, the calculation of $\langle{\bf X}\widehat{\bf Y}'\rangle_\tau$ directly follows from Equation~\ref{eq:xy}:
\begin{align}
 \langle{\bf X}\widehat{\bf Y}'\rangle_\tau = \langle u^2\rangle {\bf W}^{\tau-1}{\bf V}.
\end{align}
Our analytical solution of ${\bf W}_{out}$ assumes the true variance of the input. By appeal to the central limit theorem, we can expect that if one calculates the memory curve for a given ESN numerically, due to finite training size, the results should vary according to a normal distribution around the analytical values. We should note that  previous attempts to characterize the memory curve of ESN \cite{PhysRevLett.92.148102} used an annealed approximation over the Gaussian Orthogonal Ensemble (GOE) to simplify the problem. Here, in contrast, we presented an exact solution to this problem.

\section{Results}
\label{sect:result}
In this section we calculate the memory capacity of a given ESN using our analytical solution and compare it with  numerical estimations. For the purpose of demonstration, we create ten $N\times N$ reservoir weight matrices ${\bf W}$ and corresponding $N\times 1$ input weight matrix  ${\bf V}$ by sampling a zero-mean normal distribution with standard deviation of $1$. We then rescale the weight matrix to have a spectral radius of $\lambda^*=\lambda$. For each $\{{\bf W},{\bf V}\}$ pair we run the system with an input stream of length $5000$. We discard the first $2000$ reservoir states and use the rest to calculate $MC_\tau$, and repeat this experiment $10$ times to calculate the average the $\tau$-delay memory capacity $\overline{MC}_\tau$. As we will see, the variance in our result is low enough for 10 runs to give us a reliable average behavior. We choose $1\le\tau\le 100$, and try the experiment with $N\in\{ 25, 50, 75, 100\}$ and $\lambda\in\{0.1,0.50,0.95\}$.
\begin{figure}[t!]
\centering
\subfloat[distribution of ${\bf XX}'^*$]{
\includegraphics[width=2.25in]{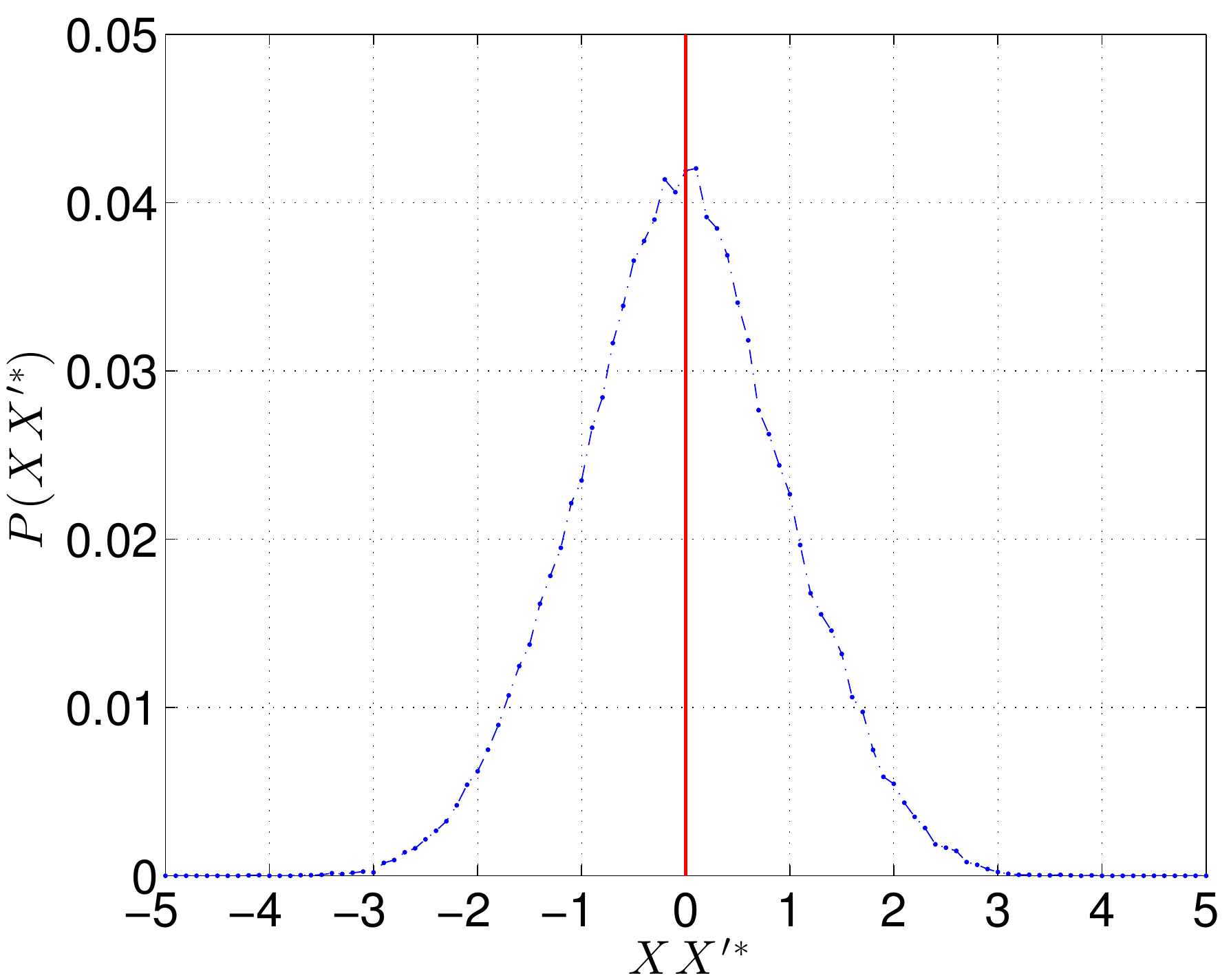}
\label{fig:xxdist}
}
\subfloat[distribution of ${\bf XY}'^*$]{
\includegraphics[width=2.25in]{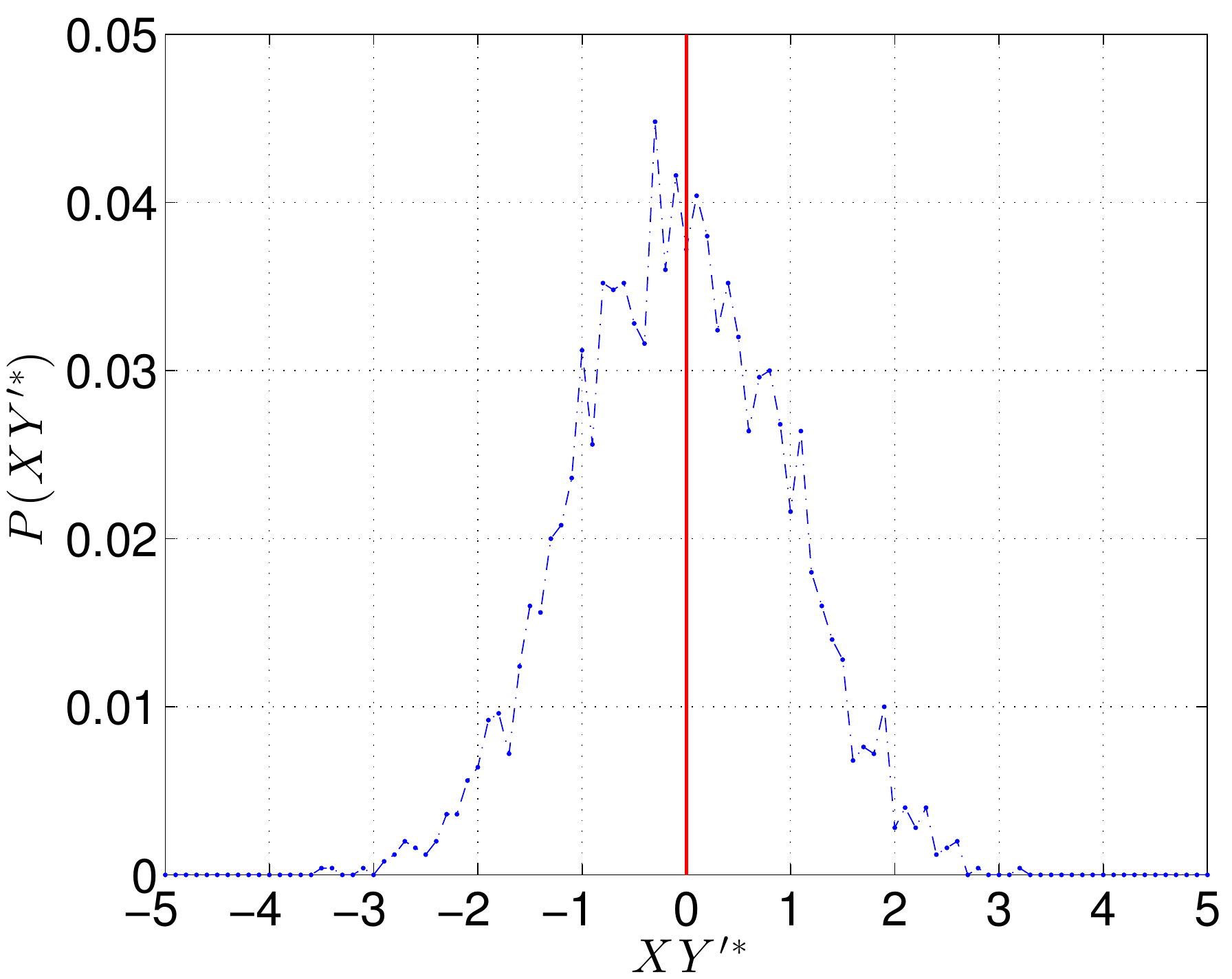}
\label{fig:xydist}
}\\
\subfloat[Memory curve $MC_\tau$]{
\includegraphics[width=2.25in]{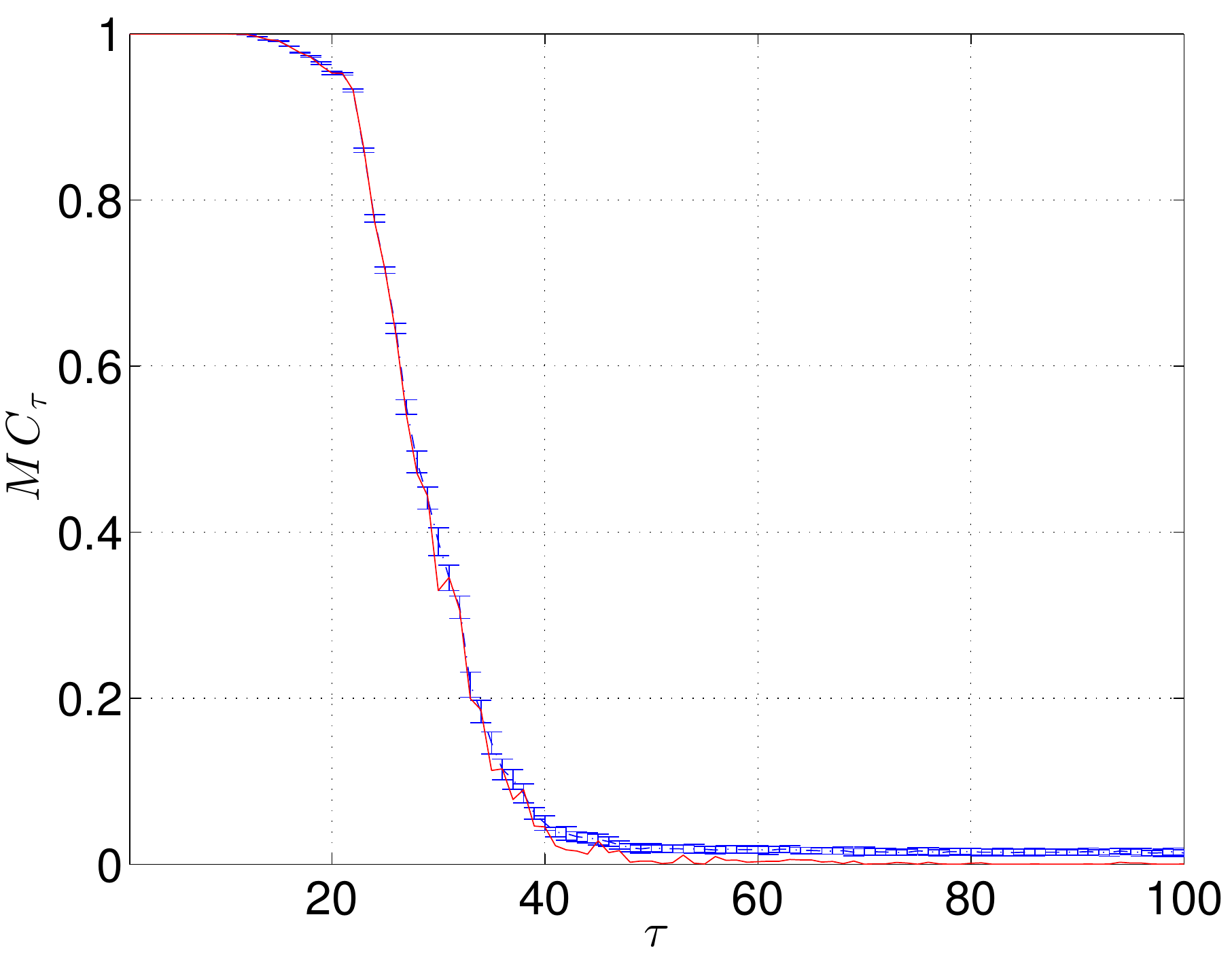}
\label{fig:mcplot}
}
\caption{Agreement between analytical and simulation results for $\langle{\bf XX'}\rangle$ (a), $\langle{\bf XY'}\rangle$ (b) and the memory curve $MC_\tau$ (c).}
\label{fig:mc_sensitivity}
\end{figure}

Figure~\ref{fig:xxdist} and Figure~\ref{fig:xydist} illustrate  the probability distribution of our simulated results for the entries of $\langle XX'\rangle$ and $\langle XY'\rangle$ for a sample ESN with $N=50$ nodes and spectral radius $\lambda=0.95$. We drove the ESN with 20 different input time series and for each input time series we calculated the matrices $\langle XX'\rangle$, $\langle X\widehat{Y}'\rangle$. To look at all the entries of $\langle XX'\rangle$ at the same time we create a dataset ${\bf XX}'^*$ by shifting and rescaling each entry of $\langle{\bf XX'}\rangle$ with the corresponding analytical values so all entries map onto a zero-mean normal distribution. As expected there is no skewness in the result, suggesting that all values follow a normal distribution centered at the analytical calculations for each entry. Similarly for Figure~\ref{fig:xydist}, we create a dataset ${\bf XY}'^*$ by shifting and rescaling each entry of $\langle{\bf XY'}\rangle$ with the corresponding analytical values to observe that all values follow a normal distribution centered at the analytical values with no skewness.

Figure~\ref{fig:mcplot} shows the complete memory curve $MC_\tau$ for the sample ESN. Our analytical results (solid red line) are in good agreement with simulation results (solid blue line). Note that our results are exact calculations and not approximation, therefore the analytical $MC_\tau$ curve also replicates the fluctuations for various values of $MC_\tau$ that are {\em signatures} of a particular instantiation of the ESN model.
\begin{figure}[ht!]
\centering
\subfloat[$N=25$]{
\includegraphics[width=2.25in]{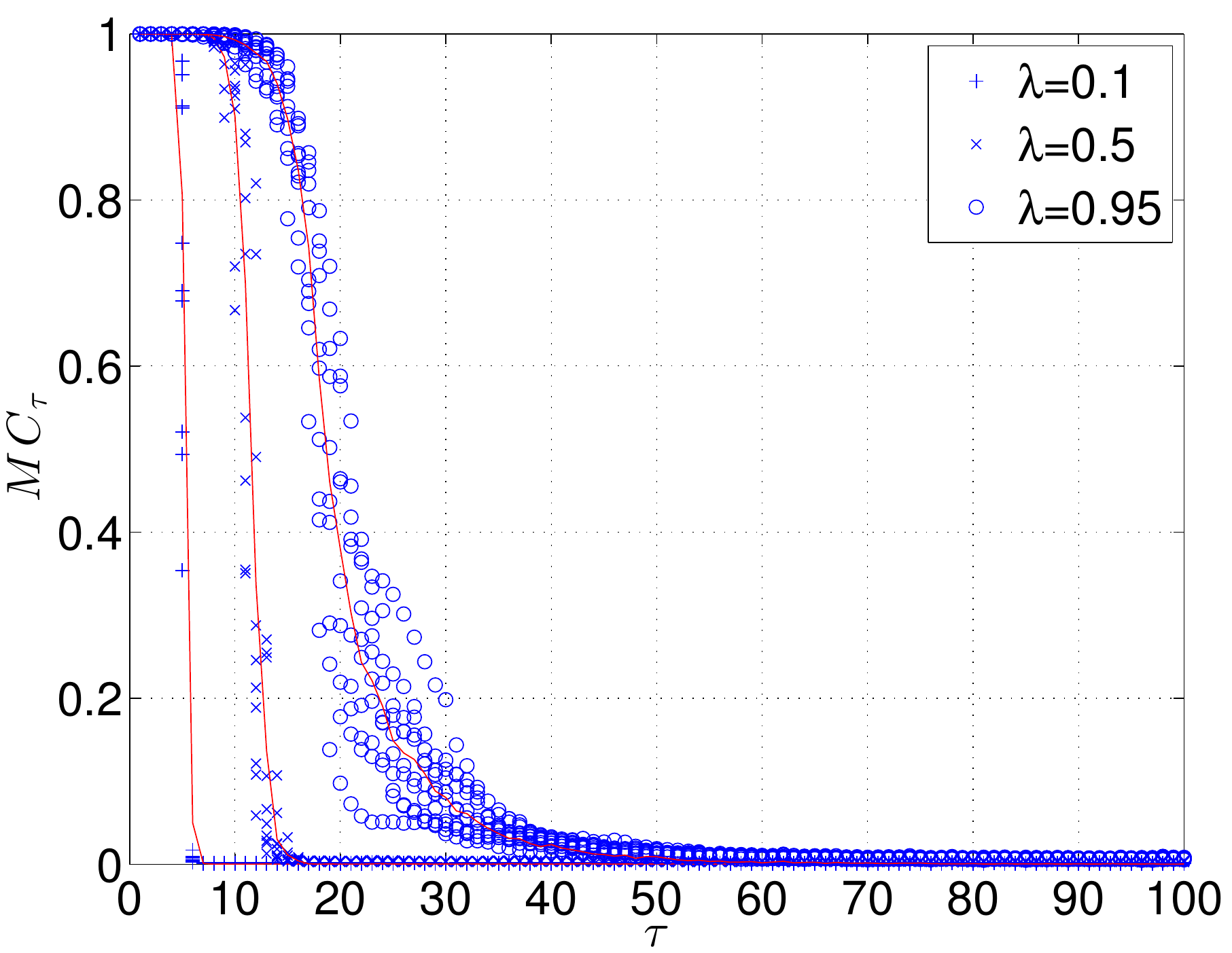}
}
\subfloat[$N=50$]{
\includegraphics[width=2.25in]{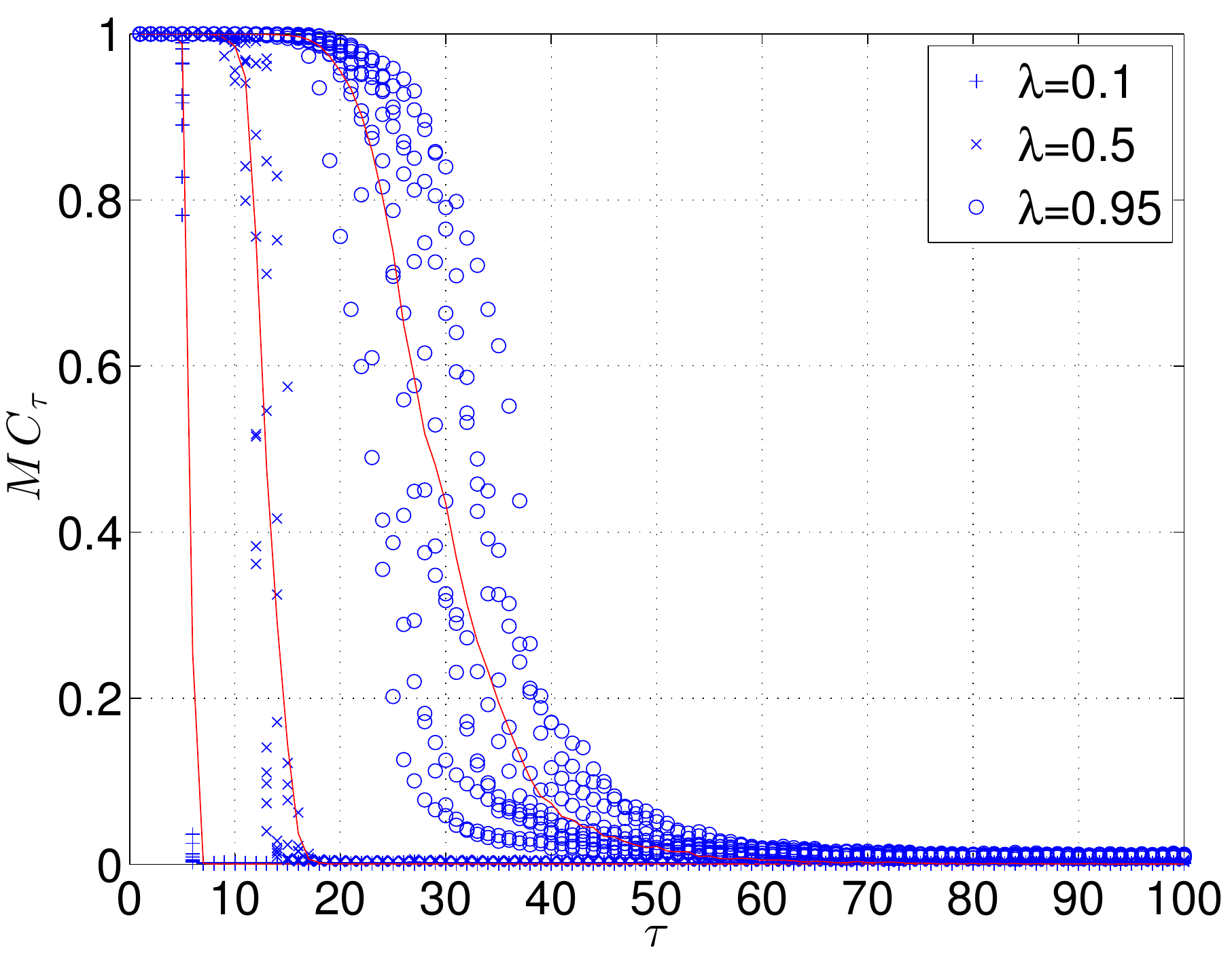}
}\\
\subfloat[$N=75$]{
\includegraphics[width=2.25in]{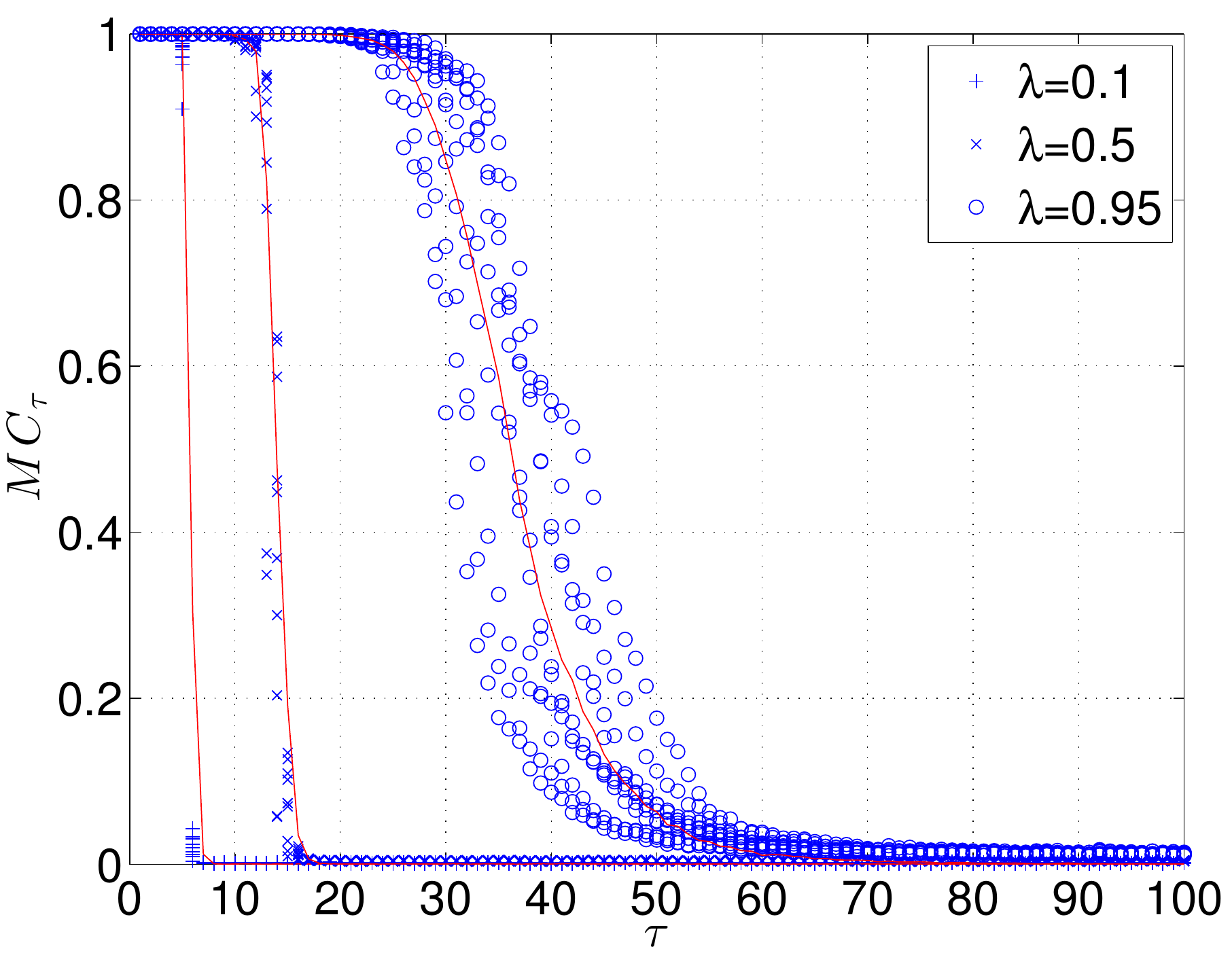}
}
\subfloat[$N=100$]{
\includegraphics[width=2.25in]{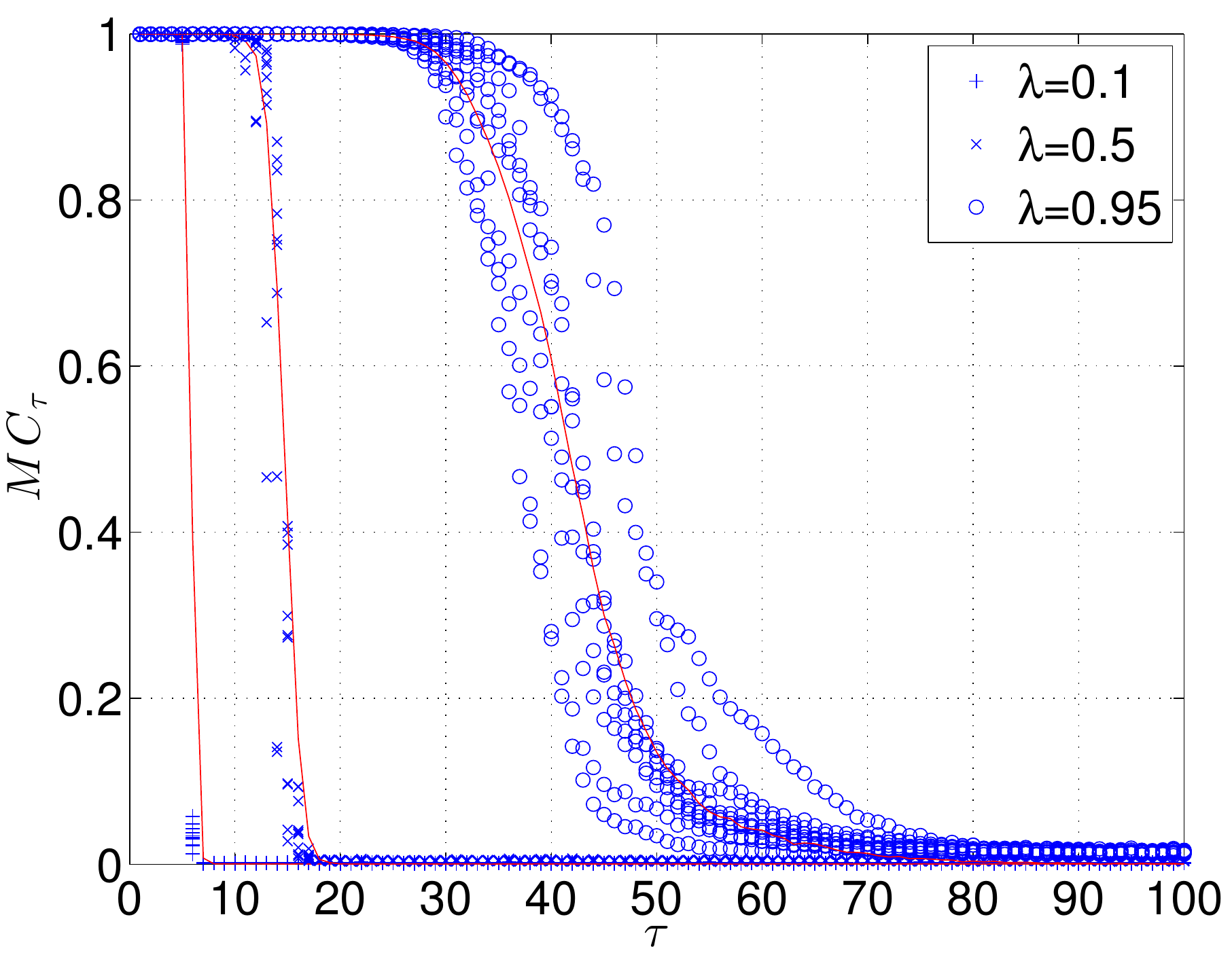}
}
\caption{Sensitivity of the analytical (red solid lines) and simulation results (data markers) for the memory curve $MC_\tau$ to changes in the system size $N$ and the spectral radius $\lambda$. The data were generated from 20 systems each driven with 20 different input streams. For all $N$ and $\lambda$ values, the analytical and simulated results are in good agreement. However, as the spectral radius approaches $\lambda=1$ the variance of the simulated results increases, suggesting that the system is approaching the chaotic dynamical phase.}
\label{fig:mc_sensitivity}
\end{figure}

Next, we analyze the accuracy of our analytical results with respect to changes in the reservoir size $N$ and its spectral radius $\lambda$.
Figure~\ref{fig:mc_sensitivity} shows the result of this analysis, and  reveals two interesting trends for accuracy and memory behavior for different $N$ and $\lambda$. For all $N$ and $\lambda$ the analytical calculation of $MC_\tau$ agrees very well with the numerical simulation. However, as we approach $\lambda=1$,  the variance in the simulation result increases during the phase transition from $MC_\tau=1$ to $MC_\tau=0$, likely because the reservoir  approaches the onset of chaos, e.g., $\lambda=1$.

The behavior of the memory function also shows interesting behavior. For small $N<50$, the transition from high to low $MC_\tau$ occurs very close to $\tau=N$, as expected from the fundamental limit $MC\le N$. However, as the reservoir size grows, the position of the transition in $MC_\tau$ diverges from $N$. Note that our analytical calculation is equivalent to using infinite size training data for calculating the output weights, therefore divergence of the actual memory capacity from the bound $N$ cannot be attributed to finite training size. Determining the reason for this discrepancy requires a more careful analysis of the memory function.

\section{Conclusion and Future Work}
\label{sect:future-work}

Our aim is to go beyond only the memory bounds in dynamical system and develop a rigorous understanding of the expected memory of the system given its structure and a desired input domain. Here, we have built a formal framework that expresses the memory as a function of the system structure and autocorrelation structure of the input. Previous attempts to characterize the memory in ESN used an annealed approximation to simplify the problem. Our approach, however, gives an exact solution for the memory curve in a given ESN. Our analytical results agree very well with numerical simulations. However, discrepancies between the analytical memory curve and the fundamental limit of memory capacity hint at a hidden process that prevents the  memory capacity from reaching its optimal value. We leave careful analysis of this deficiency for future work. In addition, we will study the presented framework to gain understanding of the effect of the structure of the system on the memory. We are currently  extending this framework to  inputs with non-uniform correlation structure. Other natural extensions are to calculate expected performance for an arbitrary output function and  analyze systems with finite dynamical range.

\subsection{Acknowledgments}
\label{sect:acks}
We thank Lance Williams, Guy Feldman, Massimo Stella, Sarah Marzen, Nix Bartten,  and Rajesh Venkatachalapathy for stimulating discussions. This material is based upon work supported by the National Science Foundation under grants CDI-1028238 and CCF-1318833.

%
\label{sect:bib}
\bibliographystyle{plain}
\bibliography{rc_symp_bica2014}

\begin{thebibliography}{10}

\bibitem{Bertschinger:2004p1450}
N.~Bertschinger and T.~Natschl{\"a}ger.
\newblock Real-time computation at the edge of chaos in recurrent neural
  networks.
\newblock {\em Neural Computation}, 16(7):1413--1436, 2004.

\bibitem{Boedecker2009}
J.~Boedecker, O.~Obst, N.~M. Mayer, and M.~Asada.
\newblock Initialization and self-organized optimization of recurrent neural
  network connectivity.
\newblock {\em HFSP Journal}, 3(5):340--349, 2009.

\bibitem{4905041020100501}
L.~B{\"u}sing, B.~Schrauwen, and R.~Legenstein.
\newblock Connectivity, dynamics, and memory in reservoir computing with binary
  and analog neurons.
\newblock {\em Neural Computation}, 22(5):1272--1311, 2010.

\bibitem{Dambre:2012fk}
J.~Dambre, D.~Verstraeten, B.~Schrauwen, and S.~Massar.
\newblock Information processing capacity of dynamical systems.
\newblock {\em Sci. Rep.}, 2, 07 2012.

\bibitem{Ganguli02122008}
S.~Ganguli, D.~Huh, and H.~Sompolinsky.
\newblock Memory traces in dynamical systems.
\newblock {\em Proceedings of the National Academy of Sciences},
  105(48):18970--18975, 2008.

\bibitem{Hermans:2011fk}
M.~Hermans and B.~Schrauwen.
\newblock Recurrent kernel machines: Computing with infinite echo state
  networks.
\newblock {\em Neural Computation}, 24(1):104--133, 2013/11/22 2011.

\bibitem{Jaeger:2001p1446}
H.~Jaeger.
\newblock Short term memory in echo state networks.
\newblock Technical Report GMD Report 152, GMD-Forschungszentrum
  Informationstechnik, 2002.

\bibitem{Jaeger:2002p1445}
H.~Jaeger.
\newblock {Tutorial on training recurrent neural networks, covering BPPT, RTRL,
  EKF and the ‚``echo state network‚" approach}.
\newblock Technical Report GMD Report 159, German National Research Center for
  Information Technology, St. Augustin-Germany, 2002.

\bibitem{Jaeger02042004}
H.~Jaeger and H.~Haas.
\newblock Harnessing nonlinearity: Predicting chaotic systems and saving energy
  in wireless communication.
\newblock {\em Science}, 304(5667):78--80, 2004.

\bibitem{Maass:2002p1444}
W.~Maass, T.~Natschl{\"a}ger, and H.~Markram.
\newblock Real-time computing without stable states: a new framework for neural
  computation based on perturbations.
\newblock {\em Neural Computation}, 14(11):2531--60, 2002.

\bibitem{Natschlaeger2003}
T.~Natschlaeger and W.~Maass.
\newblock Information dynamics and emergent computation in recurrent circuits
  of spiking neurons.
\newblock In S.~Thrun, L.~Saul, and B.~Schoelkopf, editors, {\em Proc. of NIPS
  2003, Advances in Neural Information Processing Systems}, volume~16, pages
  1255--1262, Cambridge, 2004. MIT Press.

\bibitem{5629375}
A.~Rodan and P.~Ti{\v n}o.
\newblock Minimum complexity echo state network.
\newblock {\em Neural Networks, IEEE Transactions on}, 22:131--144, Jan. 2011.

\bibitem{0957-4484-24-38-384004}
H.~O. Sillin, R.~Aguilera, H.~Shieh, A.~V. Avizienis, M.~Aono, A.~Z. Stieg, and
  J.~K. Gimzewski.
\newblock A theoretical and experimental study of neuromorphic atomic switch
  networks for reservoir computing.
\newblock {\em Nanotechnology}, 24(38):384004, 2013.

\bibitem{Snider:2005qa}
G.~Snider.
\newblock Computing with hysteretic resistor crossbars.
\newblock {\em Appl. Phys. A}, 80:1165--1172, 2005.

\bibitem{0957-4484-18-36-365202}
G.~S. Snider.
\newblock Self-organized computation with unreliable, memristive nanodevices.
\newblock {\em Nanotechnology}, 18(36):365202, 2007.

\bibitem{PhysRevE.87.042808}
D.~Snyder, A.~Goudarzi, and C.~Teuscher.
\newblock Computational capabilities of random automata networks for reservoir
  computing.
\newblock {\em Phys. Rev. E}, 87:042808, Apr 2013.

\bibitem{PhysRevLett.92.148102}
O.~L. White, Daniel~D. Lee, and H.~Sompolinsky.
\newblock Short-term memory in orthogonal neural networks.
\newblock {\em Phys. Rev. Lett.}, 92:148102, Apr 2004.

\bibitem{DBLP:journals/nn/YildizJK12}
I.~B. Yildiz, H.~Jaeger, and S.~J. Kiebel.
\newblock Re-visiting the echo state property.
\newblock {\em Neural Networks}, 35:1--9, 2012.

\end{thebibliography}


\end{document}